\documentclass[letterpaper, 10pt, conference]{ieeeconf}  

\IEEEoverridecommandlockouts                              

\usepackage{graphics} 
\usepackage{amsmath} 
\usepackage{amssymb}  
\usepackage{amsfonts}
\usepackage{xcolor}
\usepackage{subfigure}

\usepackage[linesnumbered,ruled,vlined]{algorithm2e}

\usepackage{listings}

\usepackage{graphicx}
\usepackage{booktabs}
\usepackage{multirow}
\usepackage{multicol}
\usepackage{bm}
\usepackage{threeparttable}
\usepackage{enumerate}
\usepackage{tabularx}

\usepackage{cite}

\makeatletter
\let\NAT@parse\undefined
\makeatother
\usepackage{hyperref}
\hypersetup{ 
    colorlinks=true, 
    urlcolor=blue, 
    linkcolor=blue, 
    citecolor=green 
}

\title{\LARGE \bf
OregairuChar: A Benchmark Dataset for Character Appearance Frequency Analysis in My Teen Romantic Comedy SNAFU
}

\author{
Qi Sun, Dingju Zhou, Lina Zhang
}

\begin{document}

\maketitle

\begin{abstract}


The analysis of character appearance frequency is essential for understanding narrative structure, character prominence, and story progression in anime. In this work, we introduce OregairuChar, a benchmark dataset designed for appearance frequency analysis in the anime series My Teen Romantic Comedy SNAFU. The dataset comprises 1600 manually selected frames from the third season, annotated with 2860 bounding boxes across 11 main characters. OregairuChar captures diverse visual challenges, including occlusion, pose variation, and inter-character similarity, providing a realistic basis for appearance-based studies. To enable quantitative research, we benchmark several object detection models on the dataset and leverage their predictions for fine-grained, episode-level analysis of character presence over time. This approach reveals patterns of character prominence and their evolution within the narrative. By emphasizing appearance frequency, OregairuChar serves as a valuable resource for exploring computational narrative dynamics and character-centric storytelling in stylized media.

\end{abstract}


\section{Introduction \label{sec:introduction}}

Recognizing and quantifying character presence over time is a crucial step toward understanding narrative structure, character prominence, and interpersonal dynamics in episodic visual media such as anime~\cite{kim2022animeceleb}. In particular, analyzing how frequently and when specific characters appear can offer valuable insights into story pacing, emotional arcs, and thematic emphasis. Despite its relevance, fine-grained appearance frequency analysis remains underexplored in the anime domain, largely due to the lack of high-quality, character-level annotated datasets. Moreover, the stylized nature of anime, characterized by diverse artistic representations, exaggerated expressions, and frequent occlusions, poses significant challenges for conventional object detection pipelines trained on natural image datasets such as COCO~\cite{lin2014microsoft} or Pascal VOC~\cite{everingham2010pascal}. These challenges underscore the need for dedicated datasets and robust object detection models specifically designed for character-centric temporal analysis in stylized media~\cite{zheng2020cartoon, li2022graph, qi2024bridge}.

Recent advances in anime character understanding have led to the development of various datasets tailored for tasks such as character classification, facial recognition, or retrieval~\cite{chao2019animeface, branwen2019danbooru}. Broadly, these datasets fall into two categories. The first focuses on facial or portrait-level analysis using cropped still images. For example, the Anime Face Dataset~\cite{chao2019animeface} and Danbooru2019~\cite{branwen2019danbooru} contain large volumes of character portraits or facial attributes, enabling low-level tasks like identity classification or clustering. However, they lack temporal continuity, consistent character identity tracking across frames, and scene-level annotations, making them unsuitable for modeling character appearance dynamics over time. The second category targets more contextual and narrative-rich content, such as manga or selected video clips. Datasets like Manga109~\cite{matsui2017sketch} and DAF:re~\cite{rios2021dafre} incorporate a broader visual context, yet suffer from limited annotation granularity, restricted style or genre coverage, and lack explicit temporal frequency labels. As a result, existing resources fall short of supporting fine-grained, longitudinal character appearance frequency analysis, which is essential for understanding narrative emphasis, character centrality, and evolving interpersonal relationships in anime.

These limitations highlight a persistent bottleneck in the field: the lack of a dedicated, large-scale dataset that enables accurate, temporally consistent character detection across entire anime series. In the absence of such a benchmark, it remains challenging to develop and evaluate models for long-term character tracking, appearance frequency estimation, or narrative structure analysis at scale. To facilitate progress, there is a clear need for a dataset that offers not only sufficient size and stylistic diversity, but also fine-grained character-level annotations aligned with scene and episode boundaries. Crucially, the dataset must emphasize temporal coherence, enabling consistent identity association across frames and episodes, which is essential for studying character prominence, co-occurrence patterns, and evolving narrative dynamics in serialized visual media.

To address this gap, we present OregairuChar, a high-quality dataset specifically curated for full-body anime character detection in long-form animated content. Built from the third season of My Teen Romantic Comedy SNAFU (Oregairu), the dataset features 1600 carefully selected frames annotated with 2860 bounding boxes spanning 11 main characters. The choice of series ensures a consistent visual style, a well-defined character roster, and rich narrative continuity, which makes it ideal for modeling temporally coherent character appearances. To establish baseline performance, we evaluate three representative object detection models, using standard COCO-style metrics. While YOLOv5 achieves the highest mean Average Precision (mAP), all models struggle under occlusion and inter-character similarity. This observation underscores the dataset’s complexity and its utility as a benchmark for stylized detection tasks. Beyond benchmarking, we demonstrate the practical value of accurate character detection by conducting an automated character appearance frequency analysis, which enables data-driven insights into narrative structure and character prominence over time. By releasing the dataset and accompanying tools, we aim to foster further research in stylized visual understanding, temporal reasoning, and anime-focused narrative analysis.

The contributions of this work are summarized as follows:
\begin{enumerate}
    \item We present OregairuChar, a large-scale, high-quality dataset for anime character detection, featuring over 2860 annotated instances across 1600 frames from the My Teen Romantic Comedy SNAFU series.
    \item We provide a comprehensive evaluation of object detection models on stylized anime content, revealing performance gaps and domain-specific challenges.
    \item We demonstrate a novel application of anime character detection by conducting automated character appearance frequency analysis throughout the series.
\end{enumerate}

\section{Related Work}

\subsection{General Object Detection Benchmarks and Models}
Object detection has experienced significant advancements with the advent of deep learning, particularly through the development of large-scale benchmarks such as PASCAL VOC~\cite{everingham2010pascal} and MS COCO~\cite{lin2014microsoft}. Models like Faster R-CNN~\cite{ren2015faster}, YOLO~\cite{redmon2016you}, and the more recent transformer-based DETR~\cite{carion2020end} and DINO~\cite{zhang2022dino} have become standard baselines for detection tasks. These methods demonstrate impressive performance on natural image datasets, benefiting from abundant data and extensive fine-tuning. However, these detectors often assume domain characteristics typical of real-world images, including consistent lighting, realistic textures, and well-structured object boundaries. When applied to stylized domains such as anime, where visual abstraction and exaggeration are common, the performance of these models deteriorates significantly.

\subsection{Stylized and Domain-Specific Detection}
Stylized object detection, such as in anime, manga, or game scenes, introduces unique challenges due to abstract visual elements, non-standard textures, and frequent occlusions. Conventional detectors, even when pre-trained on large datasets, struggle with such domain shifts. Research efforts have explored stylized domains like cartoons~\cite{zheng2020cartoon} and comics~\cite{nguyen2017comic}, revealing the need for tailored datasets and specialized models. Several domain adaptation methods have attempted to bridge the gap between real and stylized domains using synthetic-to-real transfer learning or adversarial training. Notably, Tzeng et al.~\cite{tzeng2017adversarial} proposed adversarial discriminative domain adaptation, and Chen et al.~\cite{chen2018domain} introduced Domain Adaptive Faster R-CNN for object detection in the wild. However, in anime detection, such methods remain underdeveloped, partly due to the lack of high-quality, large-scale benchmarks that reflect real-world anime content and character variability.

\subsection{Datasets for Anime and Non-Photorealistic Content}
Several datasets have been introduced for anime or animation-related tasks. Notable among them are Danbooru2020~\cite{branwen2019danbooru}, a large-scale anime image dataset with tag annotations, and AnimeFaceDataset~\cite{chao2019animeface}, a face detection dataset for anime-style characters. While useful for classification or face detection, these datasets are either weakly labeled or limited to constrained tasks, lacking detailed bounding box annotations and multi-character scenes. Another related effort is Manga109~\cite{matsui2017sketch}, a comic-based dataset annotated with panels, faces, and text regions. Although useful for structural layout understanding, it is limited in scope and does not target character-level detection. To date, few datasets provide dense, frame-level annotations of characters within full-length anime series, which hinders progress on complex detection tasks involving interactions, occlusions, and pose variability. In this context, OregairuChar fills an important gap by offering a high-quality, densely annotated, and temporally consistent dataset focused on multi-character anime scenes. It serves as both a benchmark for detection models and a testbed for stylized visual understanding.

\section{Dataset Construction and Analysis}

\subsection{Dataset Annotation}

To facilitate automated analysis of character appearance frequency in anime, we construct a new benchmark dataset named OregairuChar, curated from the third season of the anime series My Teen Romantic Comedy SNAFU. The dataset comprises 1600 key frames, manually selected to ensure a balanced and representative sampling of scenes across diverse narrative contexts, including classroom interactions, outdoor sequences, and emotionally charged dialogues. We adopt a semi-manual annotation pipeline to achieve both accuracy and scalability. 
\begin{figure}[htbp]
    \centering
    \includegraphics[width=0.5\textwidth]{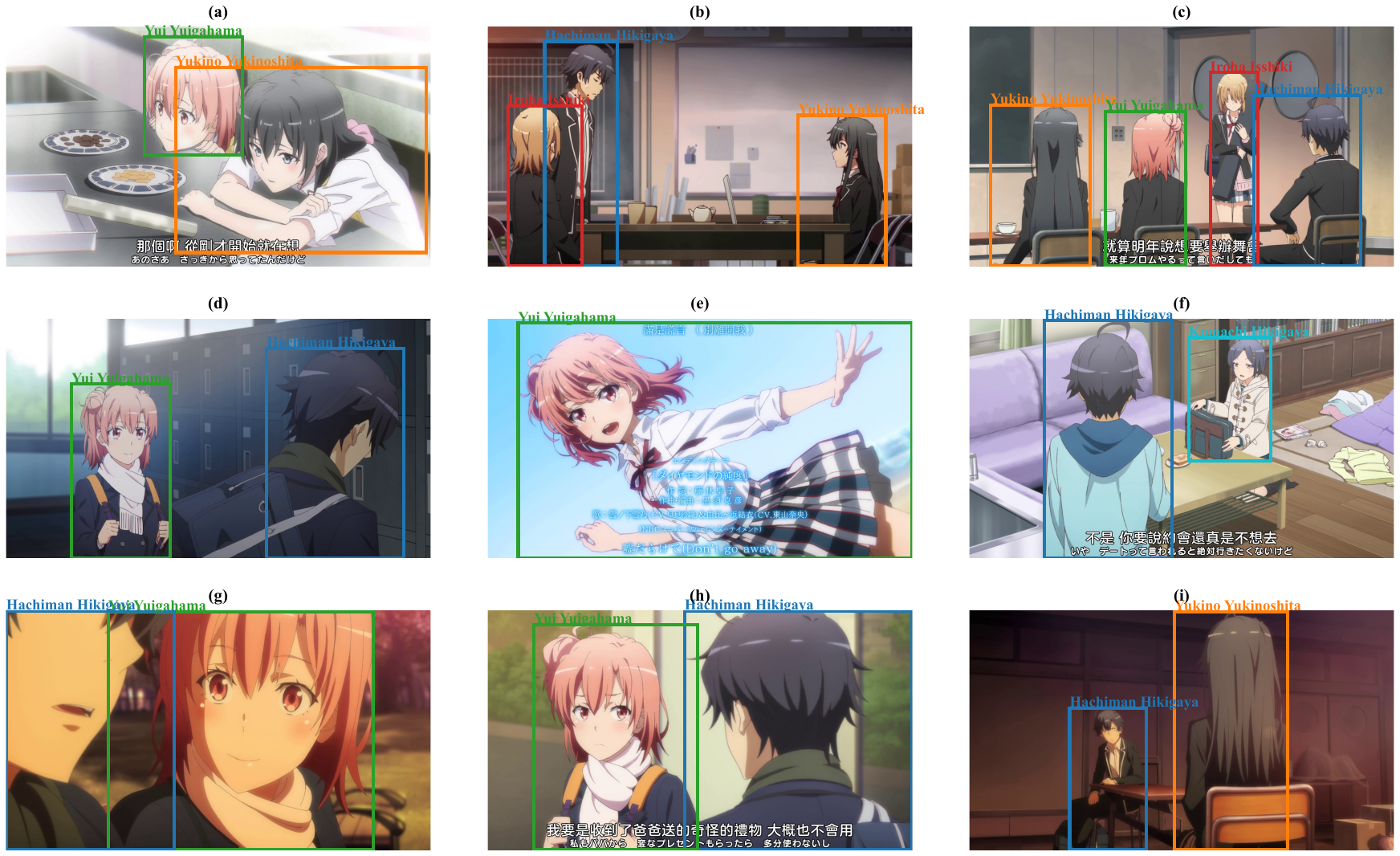}
    \caption{\textbf{Example annotation in OregairuChar.} A sample frame from My Teen Romantic Comedy SNAFU with bounding boxes for multiple main characters.}
    \label{fig:example}
\end{figure}
As shown in Fig.~\ref{fig:example}, raw frames are extracted from the official Blu-ray release at a uniform sampling rate of one frame per second. Annotators are provided with detailed character design references to ensure consistent identity assignment across scenes and episodes. Bounding boxes are manually drawn around each visible main character, regardless of variations in pose, scale, or partial occlusion. Each bounding box is labeled with the character's identity from a predefined list of 11 principal characters. All annotations undergo a two-stage quality control process: initial labeling is performed independently by three annotators, followed by cross-validation and correction by senior reviewers with domain expertise. This workflow yields a high-quality, densely annotated dataset that supports training and evaluation of object detection models in stylized animation domains.

\subsection{Dataset Statistics}

The finalized OregairuChar dataset contains 1600 fully annotated key frames, yielding a total of 2860 bounding boxes across 11 principal characters. Each character appears with sufficient frequency to facilitate reliable supervised learning. 
\begin{figure}[htbp]
    \centering
    \includegraphics[width=0.5\textwidth]{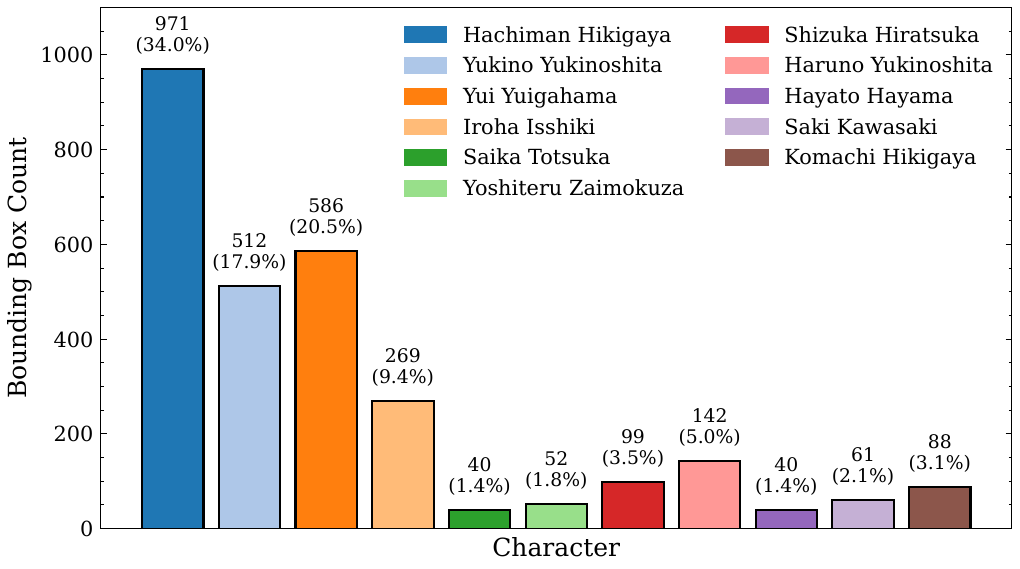}
    \caption{\textbf{Bounding‐box counts per character in OregairuChar.} The plot highlights the protagonist Hachiman Hikigaya’s dominance and the long-tail presence of supporting roles.}
    \label{fig:distribution}
\end{figure}
As shown in Fig.~\ref{fig:distribution}, the distribution of bounding boxes per character exhibits the inherent narrative imbalance typical of anime series. 
For example, the protagonist Hachiman Hikigaya dominates the visual presence due to his central narrative role, while secondary characters such as Yoshiteru Zaimokuza or Hayato Hayama appear less frequently and more episodically. Compared with other characters, Yukino Yukinoshita, Yui Yuigahama, and Iroha Isshiki appear with higher frequency, which is consistent with their roles as primary characters in the anime. For instance, the differing appearance frequencies among main and supporting characters highlight the narrative-driven imbalance in character visibility. These statistics indicate a significant level of visual diversity and challenge, which are essential for evaluating the robustness of object detection models in stylized domains.

Furthermore, both lighting conditions and background textures vary across different scene contexts, including classroom interiors, nighttime streets, and natural outdoor environments. We further compute the average bounding box size and aspect ratio for each character class to inform anchor box design and scale normalization strategies in modern detection frameworks such as YOLOv5 and Faster R-CNN. As summarized in Table~\ref{tab:bbox_stats}, characters differ significantly in terms of spatial footprint, reflecting differences in narrative prominence, typical framing, and physical scale. For example, close-up shots of Shizuka Hiratsuka yield larger bounding boxes on average, while background characters such as Saika Totsuka are associated with smaller, less prominent regions. These statistics provide actionable priors for domain-specific model tuning in anime-style object detection tasks.

\begin{table}[htbp]
    \centering
    \caption{Per-character bounding box statistics in OregairuChar}
    \label{tab:bbox_stats}
    \begin{tabular}{lccc}
        \toprule
        \textbf{Character} & \textbf{Avg. Width} & \textbf{Avg. Height} & \textbf{Aspect Ratio} \\
        \midrule
        Hachiman Hikigaya & 639.71 & 917.47 & 1.80 \\
        Yukino Yukinoshita & 637.83 & 873.79 & 1.73 \\
        Yui Yuigahama & 564.10 & 860.98 & 1.92 \\
        Iroha Isshiki & 537.83 & 850.80 & 1.95 \\
        Komachi Hikigaya & 524.65 & 822.34 & 1.89 \\
        Hayato Hayama & 572.20 & 904.03 & 2.14 \\
        Shizuka Hiratsuka & 689.47 & 932.45 & 1.70 \\
        Haruno Yukinoshita & 586.52 & 864.90 & 1.76 \\
        Saika Totsuka & 463.07 & 830.27 & 2.26 \\
        Saki Kawasaki & 546.26 & 873.46 & 1.82 \\
        Yoshiteru Zaimokuza & 667.04 & 868.98 & 1.42 \\
        \bottomrule
    \end{tabular}
\end{table}

\subsection{Challenges in Character Appearance Frequency Analysis}

\begin{figure}[htbp]
    \centering
    \subfigure[Visually similar characters.]{
        \includegraphics[width=0.5\textwidth]{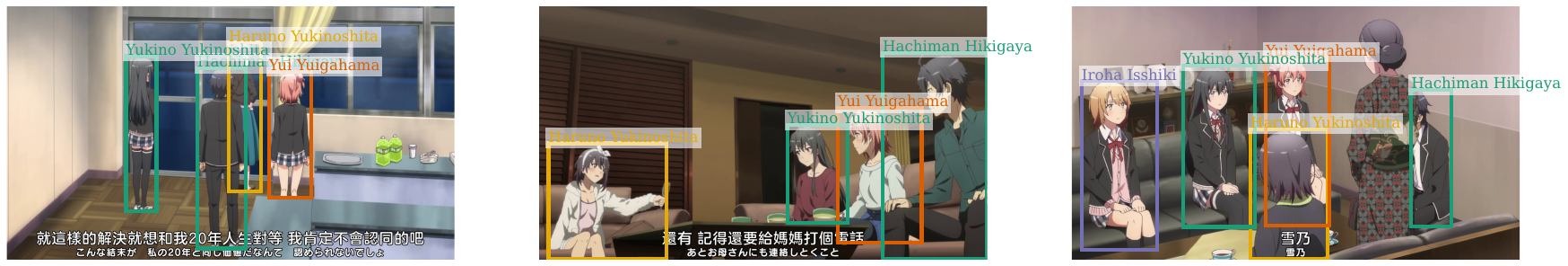}
    }
    \hfill
    \subfigure[Partial occlusions and back views.]{
        \includegraphics[width=0.5\textwidth]{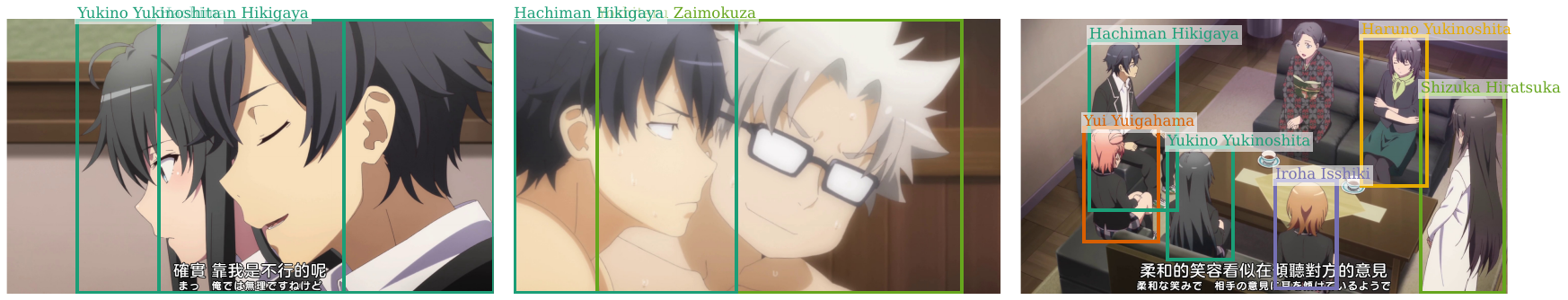}
    }
    
    \vspace{0.3cm}
    
    \subfigure[Intra-class stylistic variation.]{
        \includegraphics[width=0.5\textwidth]{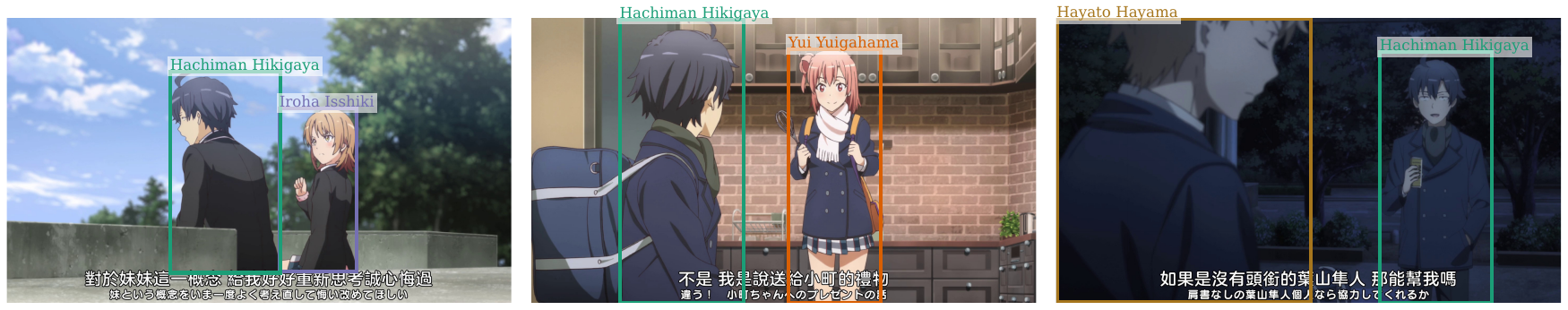}
    }
    
    \caption{\textbf{Challenges in character appearance frequency analysis illustrated by OregairuChar.}}
    \label{fig:anime_challenges}
\end{figure}

OregairuChar presents a series of unique challenges that complicate accurate character detection and downstream appearance frequency analysis in anime video content. Unlike object detection in natural scenes, anime character understanding requires fine-grained identity recognition under highly stylized and variable visual conditions. To better illustrate these domain-specific difficulties, several representative challenges observed in OregairuChar are shown in Fig.~\ref{fig:anime_challenges}.

\begin{enumerate}
    \item \textbf{High visual similarity among characters.} Many characters share nearly identical school uniforms, hairstyles, and facial features, particularly within the same grade or role. This visual homogeneity makes it difficult to reliably distinguish character identities, especially in crowded scenes or low-resolution frames.

    \item \textbf{Non-frontal views and frequent occlusions.} Characters often appear in side or back views, or are partially occluded by objects or other characters. These conditions challenge models that rely on complete facial or body features, and directly affect temporal continuity in appearance frequency estimation.

    \item \textbf{Stylistic variation within the same series.} Although all frames originate from the same anime series, significant stylistic shifts occur across different episodes and scenes—ranging from lighting and color palettes to character shading and line thickness—leading to intra-class visual inconsistencies that hinder consistent detection.

    \item \textbf{Severe class imbalance in screen presence.} The dataset exhibits a long-tailed distribution, where a few protagonists dominate most frames while others appear sparsely, as shown in Fig. \ref{fig:distribution}. This imbalance affects not only detection accuracy but also biases the appearance frequency analysis, requiring careful modeling of underrepresented characters.
    
\end{enumerate}

These challenges make OregairuChar a valuable benchmark not only for fine-grained anime character detection, but also for evaluating how well models can support downstream temporal analysis tasks such as appearance frequency tracking in stylized video domains.

\section{A Baseline Approach}

\subsection{Problem Formulation}

The goal of this study is to detect main characters in anime frames and analyze their presence over time. Given a sequence of $N$ sampled frames $\mathcal{I} = \{I_1, I_2, \ldots, I_N\}$ from an anime series, each frame $I_i$ is an RGB image of size $H \times W$. For each frame, the model predicts a set of bounding boxes $\mathcal{B}_i$ and corresponding character labels $\mathcal{C}_i$, where each box $b_{i_j} = (x, y, w, h)$ localizes a character, and $c_{i_j} \in \{1, \ldots, K\}$ indicates one of the $K = 11$ main characters. This is framed as a multi-class object detection problem in a stylized domain, challenged by occlusion, pose variation, and high visual similarity between characters. The detection results are used to track character appearances across episodes, enabling quantitative analysis of narrative structure and character prominence. The unique visual style of anime requires models to be both spatially accurate and identity-aware across diverse visual conditions.

\subsection{Model Selection}
We adopt YOLOv5 as our baseline detector due to its well-established trade-off between detection accuracy and computational efficiency. As a one-stage anchor-based model, YOLOv5 enables real-time object detection with strong performance on benchmark datasets such as MS COCO. These characteristics make it a suitable starting point for stylized character detection in anime frames, where both high throughput and precise localization are desirable. Moreover, the flexibility of the YOLOv5 framework allows for seamless customization and integration of additional modules.

\subsection{Training Objective}

The YOLOv5 model is trained to jointly optimize bounding box localization and character identity classification. The overall objective function is a weighted sum of a localization loss and a classification loss:
\[
\mathcal{L}_{\text{total}} = \lambda_{\text{loc}} \cdot \mathcal{L}_{\text{loc}} + \lambda_{\text{cls}} \cdot \mathcal{L}_{\text{cls}},
\]
where \(\lambda_{\text{loc}}\) and \(\lambda_{\text{cls}}\) are hyperparameters that balance the contribution of each component. The localization loss combines an \(\ell_1\) regression loss for bounding box coordinates and an IoU-based loss to improve overlap consistency with ground-truth regions. For classification, binary cross-entropy with label smoothing is applied to address the overconfidence problem commonly observed in imbalanced class settings. To improve generalization, we employ a suite of standard data augmentation techniques during training, including random horizontal flipping, scaling, cropping, and color jittering. Optimization is performed using the Adam optimizer with a cosine annealing learning rate schedule, which allows for a gradual reduction in learning rate and stabilizes convergence.

\subsection{Identity Refinement via Feature Embeddings}

Due to the high visual similarity among certain characters, especially those wearing similar school uniforms or sharing facial features, the base YOLOv5 model occasionally confuses character identities. To address this issue, we introduce a feature-based refinement step at inference time. Specifically, for each predicted bounding box, a fixed-size visual embedding is extracted using a pre-trained and frozen image encoder, such as ResNet-50 or CLIP ViT. These embeddings are then compared against a reference gallery constructed from annotated exemplars in the training set. Cosine similarity is used to measure identity affinity, and predictions falling below a similarity threshold are discarded. This post-processing step enhances identity consistency and filters out visually ambiguous detections without requiring re-training.

\begin{table*}[htbp]
    \centering
    \caption{Detection metrics per character across different models}
    \label{tab:per_character_model_comparison}
    \begin{tabular}{cccccccc}
    \toprule
    \textbf{Character} & \textbf{Model} & \textbf{AP@0.5} & \textbf{AP@0.75} & \textbf{mAP@[0.5:0.95]} & \textbf{Precision} & \textbf{Recall} & \textbf{F1 Score} \\
    \midrule
    \multirow{3}{*}{Hachiman Hikigaya}
        & Faster R-CNN & 78.29\% & 76.55\% & 69.37\% & 68.26\% & \textbf{100.00\%} & 81.14\% \\
        & SSD & 91.53\% & 72.89\% & 62.67\% & \textbf{99.15\%} & 91.53\% & 95.19\% \\
        & YOLOv5 & \textbf{98.94\%} & \textbf{95.20\%} & \textbf{90.01\%} & \textbf{96.30\%} & 95.54\% & \textbf{95.92\%} \\
    \midrule
    \multirow{3}{*}{Yukino Yukinoshita}
        & Faster R-CNN & 74.71\% & 74.24\% & 65.35\% & 67.48\% & \textbf{96.51\%} & 79.43\% \\
        & SSD & 73.88\% & 49.04\% & 47.73\% & 92.17\% & 73.88\% & 82.02\% \\
        & YOLOv5 & \textbf{98.20\%} & \textbf{94.54\%} & \textbf{87.51\%} & \textbf{95.89\%} & 95.35\% & \textbf{95.62\%} \\
    \midrule
    \multirow{3}{*}{Yui Yuigahama}
        & Faster R-CNN & 76.71\% & 76.59\% & 68.70\% & 56.64\% & \textbf{96.43\%} & 71.37\% \\
        & SSD & 81.42\% & 60.59\% & 53.51\% & \textbf{98.14\%} & 81.42\% & 89.00\% \\
        & YOLOv5 & \textbf{97.49\%} & \textbf{96.62\%} & \textbf{89.35\%} & 97.41\% & \textbf{96.43\%} & \textbf{96.92\%} \\
    \midrule
    \multirow{3}{*}{Iroha Isshiki}
        & Faster R-CNN & 85.11\% & 81.10\% & 66.48\% & 39.42\% & \textbf{100.00\%} & 56.55\% \\
        & SSD & 42.38\% & 32.32\% & 27.39\% & 55.43\% & 42.38\% & 48.04\% \\
        & YOLOv5 & \textbf{99.17\%} & \textbf{91.41\%} & \textbf{85.89\%} & \textbf{98.80\%} & 92.68\% & \textbf{95.64\%} \\
    \midrule
    \multirow{3}{*}{Haruno Yukinoshita}
        & Faster R-CNN & 89.51\% & 68.86\% & 67.60\% & 22.64\% & \textbf{92.31\%} & 36.36\% \\
        & SSD & 40.86\% & 33.32\% & 29.92\% & 54.25\% & 40.86\% & 46.71\% \\
        & YOLOv5 & \textbf{93.76\%} & \textbf{92.61\%} & \textbf{81.12\%} & \textbf{99.01\%} & \textbf{92.31\%} & \textbf{95.54\%} \\
    \midrule
    \multirow{3}{*}{Shizuka Hiratsuka}
        & Faster R-CNN & 79.61\% & 73.54\% & 64.24\% & 16.46\% & \textbf{100.00\%} & 28.26\% \\
        & SSD & 24.12\% & 9.82\% & 11.82\% & 44.47\% & 24.12\% & 31.28\% \\
        & YOLOv5 & \textbf{98.57\%} & \textbf{92.63\%} & \textbf{76.03\%} & \textbf{100.00\%} & 88.53\% & \textbf{93.92\%} \\
    \midrule
    \multirow{3}{*}{Komachi Hikigaya}
        & Faster R-CNN & 75.26\% & 58.41\% & 56.79\% & 17.14\% & \textbf{92.31\%} & 28.92\% \\
        & SSD & 23.54\% & 19.61\% & 16.88\% & 44.12\% & 23.54\% & 30.70\% \\
        & YOLOv5 & \textbf{82.70\%} & \textbf{74.59\%} & \textbf{70.65\%} & \textbf{100.00\%} & 73.97\% & \textbf{85.03\%} \\
    \midrule
    \multirow{3}{*}{Yoshiteru Zaimokuza}
        & Faster R-CNN & 79.80\% & 58.12\% & 57.21\% & 18.42\% & \textbf{100.00\%} & 31.11\% \\
        & SSD & 53.26\% & 37.08\% & 36.95\% & 85.98\% & 53.26\% & 65.78\% \\
        & YOLOv5 & \textbf{99.50\%} & \textbf{86.84\%} & \textbf{85.08\%} & \textbf{96.97\%} & \textbf{100.00\%} & \textbf{98.46\%} \\
    \midrule
    \multirow{3}{*}{Saki Kawasaki}
        & Faster R-CNN & 69.09\% & \textbf{69.09\%} & 59.64\% & 20.00\% & \textbf{80.00\%} & 32.00\% \\
        & SSD & 48.72\% & 48.72\% & 40.67\% & 59.23\% & 48.72\% & 53.47\% \\
        & YOLOv5 & \textbf{79.64\%} & 59.70\% & \textbf{66.35\%} & \textbf{92.83\%} & \textbf{80.00\%} & \textbf{85.94\%} \\
    \midrule
    \multirow{3}{*}{Saika Totsuka}
        & Faster R-CNN & 74.79\% & 74.79\% & 66.54\% & 29.41\% & \textbf{100.00\%} & 45.45\% \\
        & SSD & 12.31\% &  3.55\% &  4.25\% & 18.46\% & 12.31\% & 14.77\% \\
        & YOLOv5 & \textbf{97.83\%} & \textbf{97.83\%} & \textbf{85.26\%} & \textbf{97.40\%} & 90.00\% & \textbf{93.55\%} \\
    \midrule
    \multirow{3}{*}{Hayato Hayama}
        & Faster R-CNN & 85.65\% & \textbf{74.78\%} & 66.38\% & 20.51\% & \textbf{100.00\%} & 34.04\% \\
        & SSD &  0.60\% &  0.15\% &  0.33\% &  0.90\% &  0.60\% &  0.72\% \\
        & YOLOv5 & \textbf{86.78\%} & 74.68\% & \textbf{71.21\%} & \textbf{84.93\%} & 87.50\% & \textbf{86.20\%} \\
    \bottomrule
    \end{tabular}
\end{table*}

\section{Experiments}

\subsection{Experimental Setup}
To evaluate our anime character detection framework, we conduct experiments on the proposed OregairuChar dataset. The dataset is split into 70\% training, 15\% validation, and 15\% testing, with class distribution preserved across episodes to minimize sampling bias. Training is performed with PyTorch on an NVIDIA RTX 4060Ti GPU using a batch size of 16, the Adam optimizer, and a cosine learning rate schedule, with early stopping based on validation performance. We train for up to 100 epochs. To enhance identity classification, we add an embedding-based verification step: for each predicted region, its embedding is compared to a class-specific memory bank using cosine similarity, filtering out ambiguous predictions and improving recall without compromising accuracy.

\subsection{Evaluation Metrics}

Model performance is evaluated using standard COCO-style detection metrics, including AP@0.5, AP@0.75, and mAP averaged over IoU thresholds from 0.5 to 0.95. These metrics offer a thorough assessment of localization accuracy under different overlap requirements. In addition, we report precision, recall, and F1 score to evaluate the model’s classification performance. Precision reflects the proportion of correctly predicted positives, recall measures the ability to retrieve all relevant instances, and the F1 score provides a balanced summary of both.

\subsection{Experimental Results}

To better understand per-character detection performance, we evaluate three representative models across different characters, as summarized in Table~\ref{tab:per_character_model_comparison}. Results show clear variation depending on character frequency and visual distinctiveness. Main characters such as Yukino Yukinoshita, Yui Yuigahama, and Hachiman Hikigaya achieve strong results with YOLOv5, reaching mAP values above 87\% and precision exceeding 95\%. By contrast, less prominent characters like Saika Totsuka or Hayato Hayama perform much worse under Faster R-CNN and SSD, where recall often falls below 50\%. Visually similar characters, such as Haruno Yukinoshita and Shizuka Hiratsuka, also show inconsistencies, with YOLOv5 performing well but Faster R-CNN producing low precision. Overall, these findings suggest that character frequency and visual uniqueness play a critical role in detection performance, and addressing these challenges may require targeted augmentation or improved feature discrimination.

\begin{figure*}[htbp]
    \centering
    \includegraphics[width=0.94\textwidth]{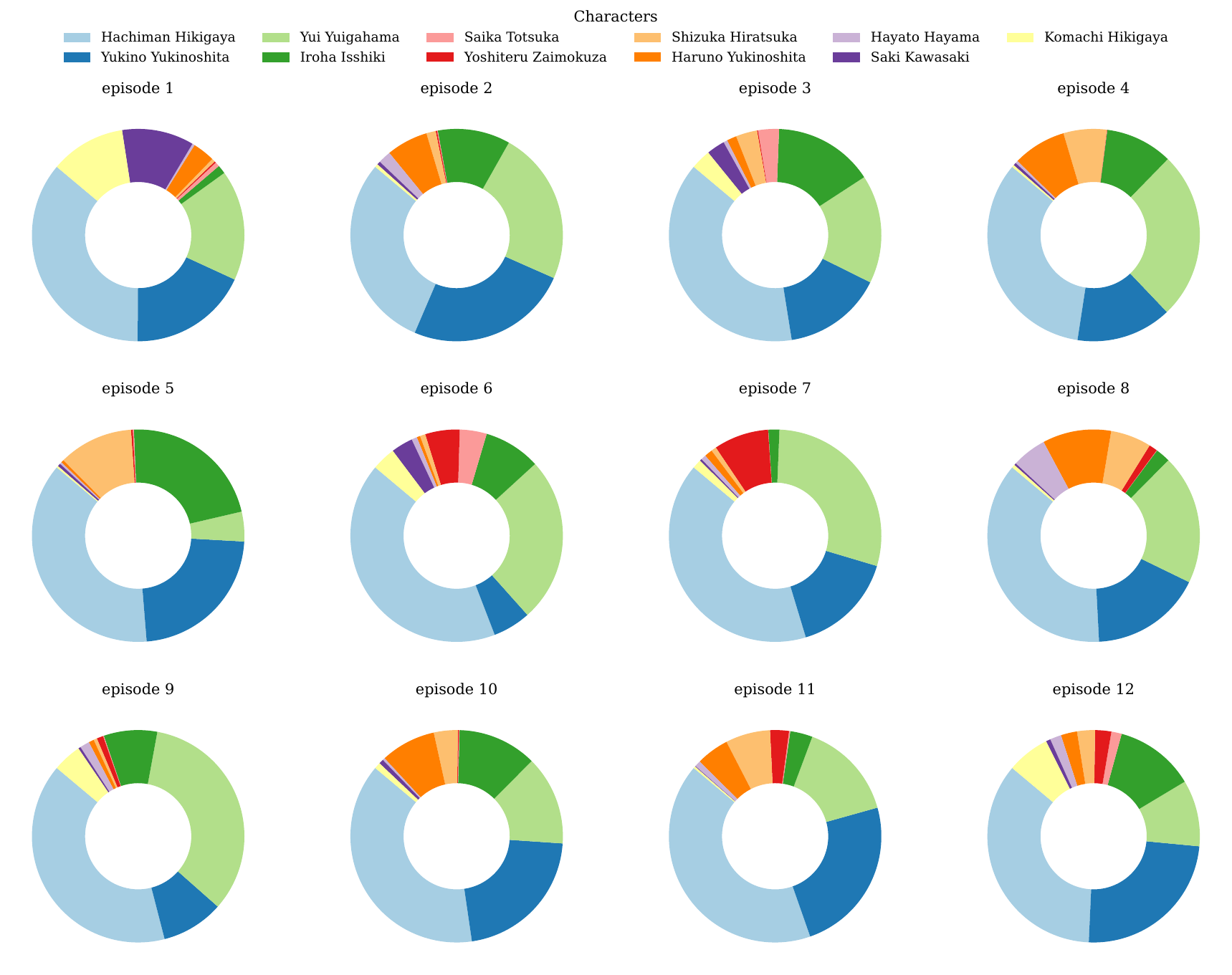} 
    \caption{Character appearance frequency and relative presence per episode in the dataset, as detected by the YOLOv5-based character detector. The plots illustrate how main characters maintain consistent prominence across episodes, while supporting characters show more sporadic and limited appearances, reflecting their narrative significance and posing challenges for detection model training under class imbalance.}
    \label{fig:character_distribution}
\end{figure*}

To further analyze character presence and distribution within the dataset, we utilize the YOLOv5-based character detector to quantify the appearance frequency of each character across different episodes. For each episode, we calculate both the absolute number of appearances and the relative proportion that each character contributes within that episode. The results, visualized in Fig.~\ref{fig:character_distribution}, reveal substantial variation in character prominence over time. Main characters, such as Hachiman Hikigaya, Yukino Yukinoshita and Yui Yuigahama, maintain a consistently high presence throughout the series, reflecting their central narrative roles. In contrast, supporting characters tend to appear sporadically and occupy only a small fraction of the total detections within any given episode. This uneven distribution poses challenges for training, particularly for rare characters whose limited presence may lead to underrepresentation during learning. These observations highlight the importance of accounting for temporal character distribution when designing and evaluating detection models in episodic content.

\begin{figure}[htbp]
    \centering
    \includegraphics[width=0.45\textwidth]{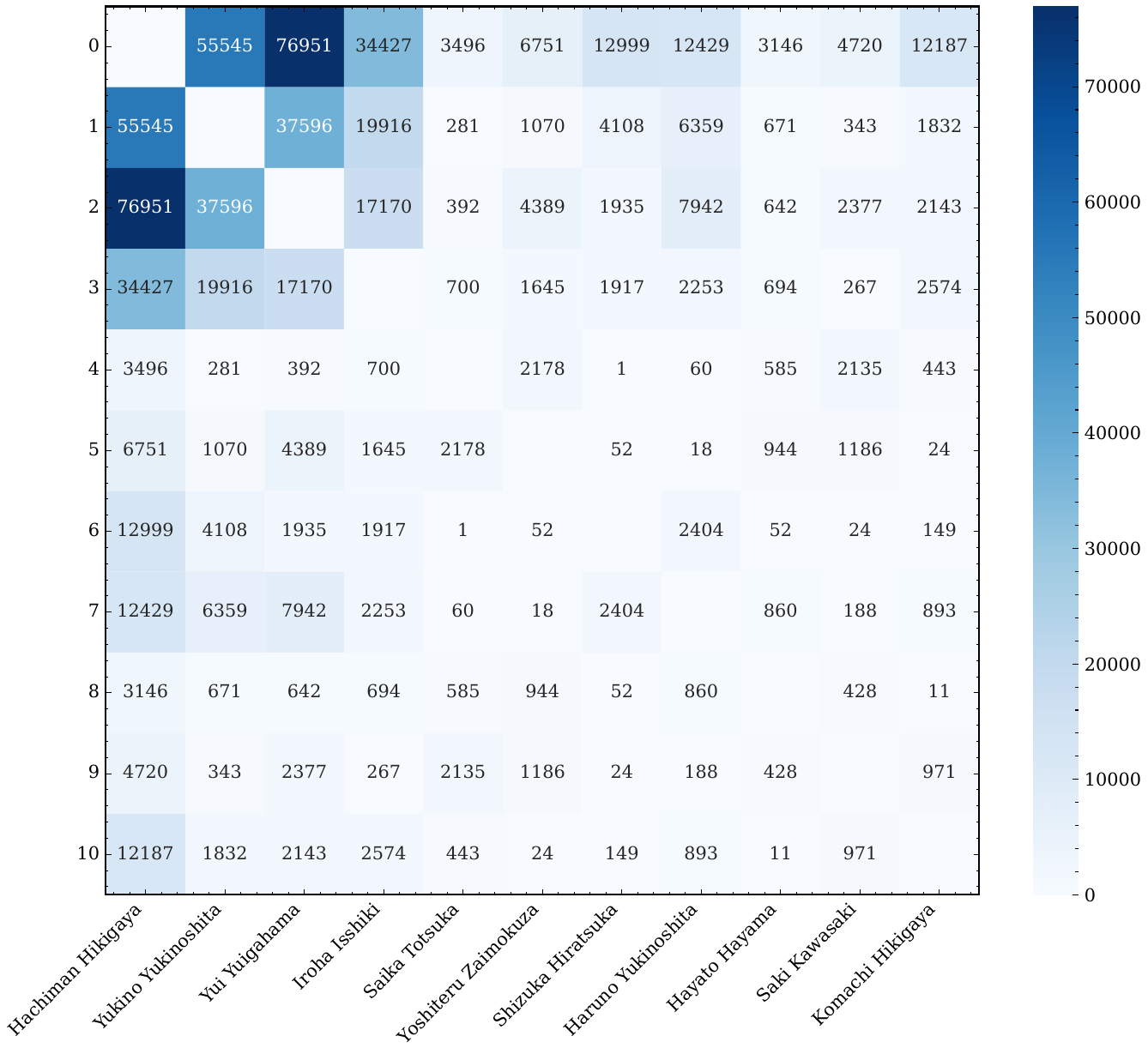}
    \caption{Character co-occurrence matrix illustrating the frequency with which pairs of characters appear together within the same frame across the analyzed video episodes.}
    \label{fig:character_co_occurrence}
\end{figure}

\section{Conclusion}

In this work, we propose OregairuChar, a benchmark dataset for analyzing character appearance frequency in the anime My Teen Romantic Comedy SNAFU. The dataset includes 1600 manually selected frames from Season 3, annotated with 2860 bounding boxes covering 11 main characters. It captures key challenges such as occlusion, pose variation, and visual similarity between characters. We evaluate several object detection models, including Faster R-CNN, YOLOv5, and SSD. The detection results are then used to analyze character appearance patterns across episodes, revealing insights into narrative focus and character dynamics. OregairuChar offers a new resource for studying anime through computer vision, supporting research on character detection and narrative analysis in stylized media. In the future, this dataset can facilitate the development of more robust models for stylized character detection and enable deeper exploration of temporal narrative patterns and character interactions.

\bibliographystyle{ieeetr}
\bibliography{references}

\end{document}